\documentstyle[a4,figfont,eepic,epsf]{article}
\newcommand{\lscott}{[ \hspace{-.5mm} [}
\newcommand{\rscott}{] \hspace{-.5mm} ]}
\newcommand{\denotation}[1]{\lscott #1 \rscott}
\newcommand{\hpsg}{\mbox{\sc hpsg}}

\begin{document}

\title{Multi-Dimensional Inheritance}

\author{Gregor Erbach\\
\thanks{This work was supported by the Deutsche Forschungsgemeinschaft,
Special Research Division 314 ``Artificial Intelligence - Knowledge-Based
Systems'' through project N3 ``Bidirectional Linguistic Deduction'' (BiLD),
the Commission of the European Communities through the project
LRE-61-061 ``Reusable Grammatical Resources,'' and Cray Systems.
Part of the work was done during a visit at the Human Communication
Research Centre, University of Edinburgh.
I would like to thank the members of the Language Technology Group at
HCRC, Chris Brew, Suresh Manandhar, Drew Moshier
and Hans Uszkoreit for their comments.
}
Universit\"at des Saarlandes \\
           FR 8.7 Computerlinguistik \\
           D-66041 Saarbr\"ucken, Germany \\
           erbach@coli.uni-sb.de\\
           \\
           \\
           (cmp-lg/9411025)}

\date{April 1994}

\maketitle

\begin{abstract}
In this paper, we present an alternative approach to multiple inheritance
for typed feature structures.
In our
approach, a feature structure can be associated with several types coming
from different hierarchies (dimensions). In case of multiple inheritance, a
type
has supertypes from different hierarchies. We contrast this approach
with approaches based on a single type hierarchy where a feature structure
has only one unique most general type, and multiple inheritance
involves computation of greatest lower bounds in the hierarchy.
The proposed approach supports current linguistic
analyses in constraint-based formalisms like HPSG, inheritance in
the lexicon, and knowledge representation for NLP systems. Finally,
we show that multi-dimensional inheritance hierarchies can be
compiled into a Prolog term representation, which allows to compute
the conjunction of two types efficiently by Prolog term unification.
\end{abstract}

\section{Introduction}

\begin{figure}[htb]
\begin{center}
\mbox{
\setlength{\unitlength}{0.0087in}
\begin{picture}(497,222)(0,-10)
\path(35,65)(130,15)
\path(380,65)(130,15)(130,15)
\path(125,65)(295,15)
\path(460,65)(295,15)(295,15)
\put(100,0){\makebox(0,0)[lb]{\smash{{{\SetFigFont{8}{9.6}{rm}Su-Wh-Rel}}}}}
\put(275,0){\makebox(0,0)[lb]{\smash{{{\SetFigFont{8}{9.6}{rm}That-Rel}}}}}
\path(85,185)(300,185)(325,185)
\path(85,163)(51,138)
\path(85,163)(131,138)(131,138)
\path(131,122)(40,86)
\path(131,122)(85,86)
\path(131,122)(125,86)
\path(131,122)(171,86)
\path(335,163)(251,128)
\path(335,163)(335,80)
\path(335,163)(425,128)
\path(251,112)(235,80)
\path(251,112)(280,80)
\path(431,112)(400,86)
\path(431,112)(475,86)
\put(190,195){\makebox(0,0)[lb]{\smash{{{\SetFigFont{8}{9.6}{rm}Sign}}}}}
\put(105,128){\makebox(0,0)[lb]{\smash{{{\SetFigFont{8}{9.6}{rm}Headed-Ph}}}}}
\put(71,170){\makebox(0,0)[lb]{\smash{{{\SetFigFont{8}{9.6}{rm}Phrase}}}}}
\put(20,70){\makebox(0,0)[lb]{\smash{{{\SetFigFont{8}{9.6}{rm}H-Su}}}}}
\put(0,128){\makebox(0,0)[lb]{\smash{{{\SetFigFont{8}{9.6}{rm}Non-Headed-Ph}}}}}
\put(65,70){\makebox(0,0)[lb]{\smash{{{\SetFigFont{8}{9.6}{rm}H-Co}}}}}
\put(111,70){\makebox(0,0)[lb]{\smash{{{\SetFigFont{8}{9.6}{rm}H-Mk}}}}}
\put(160,70){\makebox(0,0)[lb]{\smash{{{\SetFigFont{8}{9.6}{rm}H-Fi}}}}}
\put(271,70){\makebox(0,0)[lb]{\smash{{{\SetFigFont{8}{9.6}{rm}Y-N}}}}}
\put(245,117){\makebox(0,0)[lb]{\smash{{{\SetFigFont{8}{9.6}{rm}Int}}}}}
\put(420,117){\makebox(0,0)[lb]{\smash{{{\SetFigFont{8}{9.6}{rm}Rel}}}}}
\put(320,70){\makebox(0,0)[lb]{\smash{{{\SetFigFont{8}{9.6}{rm}Decl}}}}}
\put(365,70){\makebox(0,0)[lb]{\smash{{{\SetFigFont{8}{9.6}{rm}Wh-Rel}}}}}
\put(425,70){\makebox(0,0)[lb]{\smash{{{\SetFigFont{8}{9.6}{rm}Non-Wh-Rel}}}}}
\put(215,70){\makebox(0,0)[lb]{\smash{{{\SetFigFont{8}{9.6}{rm}Wh-Int}}}}}
\put(300,168){\makebox(0,0)[lb]{\smash{{{\SetFigFont{8}{9.6}{rm}Clause-Type}}}}}
\end{picture}
}
\end{center}
\caption{HPSG type hierarchy {\em with} multi-dimensional inheritance}
\label{hpsg}
\end{figure}
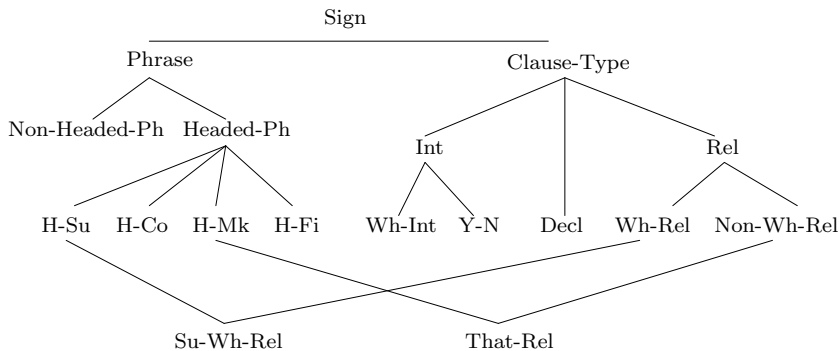

In the past years, multiple inheritance has been
increasingly used for the  description of natural languages.
Some examples are the work in Head-driven Phrase Structure Grammar
(\hpsg ) on the structure of the lexicon
\cite{Flickinger:Nerbonne:92,Pollard:Sag:87},
semantic sorts for selectional restrictions \cite{Alshawi:91},
derivational morphology
\cite{Riehemann:92,Krieger:Nerbonne:92},
and the syntax of English relative clauses \cite{Sag:94}.
As an example of this type of analysis, figure \ref{hpsg} shows
a recent \hpsg\ description of English clauses.%
\footnote{This figure is taken from a slide presented by Ivan Sag
          at the 1993 EACL meeting in Utrecht. The use of horizontal
          lines in the hierarchy to indicate a split into several
          dimensions has been adopted from the book on the Core
          Language Engine \cite{Alshawi:91}.}

In the example, relative
clauses are cross-classified according to two ``dimensions'';
on the one hand according to the Phrase-Type (Headed or Non-Headed),
and on the other hand according to the Clause-Type (Interrogative,
Declarative, or Relative). The choices within one dimension are
mutually exclusive: no structure can be described as both Headed and
Non-Headed, or as more than one of \{Int, Decl, Rel\}. However, a
structure can be assigned types from different dimensions, without
the need for a subtype that inherits from both dimensions.

In  section \ref{mdi}, we present a
concept of typing, which provides direct support for
multi-dimensional inheritance, and compare it to the type hierarchies
in Carpenter's typed feature logic.
Section \ref{mellish} shows that unification in
multi-dimensional inheritance hierarchies can be implemented
efficiently as unification of a Prolog term representation of
the types. Section \ref{winograd} applies
multi-dimensional inheritance to the problem of systemic classification.

\section{Multi-Dimensional Inheritance} \label{mdi}

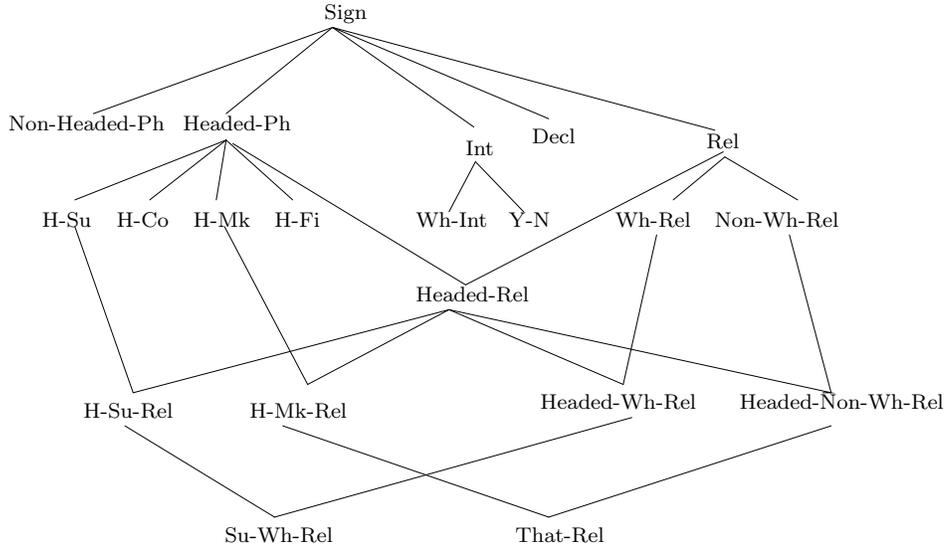
\begin{figure}[htb]
\begin{center}
\mbox{
\setlength{\unitlength}{0.0087in}
\begin{picture}(559,342)(0,-10)
\path(131,242)(40,206)
\path(131,242)(85,206)
\path(131,242)(125,206)
\path(131,242)(171,206)
\put(105,248){\makebox(0,0)[lb]{\smash{{{\SetFigFont{8}{9.6}{rm}Headed-Ph}}}}}
\put(20,190){\makebox(0,0)[lb]{\smash{{{\SetFigFont{8}{9.6}{rm}H-Su}}}}}
\put(0,248){\makebox(0,0)[lb]{\smash{{{\SetFigFont{8}{9.6}{rm}Non-Headed-Ph}}}}}
\put(65,190){\makebox(0,0)[lb]{\smash{{{\SetFigFont{8}{9.6}{rm}H-Co}}}}}
\put(111,190){\makebox(0,0)[lb]{\smash{{{\SetFigFont{8}{9.6}{rm}H-Mk}}}}}
\put(160,190){\makebox(0,0)[lb]{\smash{{{\SetFigFont{8}{9.6}{rm}H-Fi}}}}}
\put(45,75){\makebox(0,0)[lb]{\smash{{{\SetFigFont{8}{9.6}{rm}H-Su-Rel}}}}}
\put(145,75){\makebox(0,0)[lb]{\smash{{{\SetFigFont{8}{9.6}{rm}H-Mk-Rel}}}}}
\put(320,80){\makebox(0,0)[lb]{\smash{{{\SetFigFont{8}{9.6}{rm}Headed-Wh-Rel}}}}}
\put(440,80){\makebox(0,0)[lb]{\smash{{{\SetFigFont{8}{9.6}{rm}Headed-Non-Wh-Rel}}}}}
\path(195,310)(51,258)
\path(195,310)(131,258)(131,258)
\path(195,310)(280,250)
\path(195,310)(425,248)
\path(431,232)(400,206)
\path(431,232)(475,206)
\path(195,310)(325,255)
\path(281,229)(265,199)
\path(281,229)(310,199)
\path(135,240)(275,155)
\path(430,235)(275,155)
\path(40,190)(75,90)
\path(130,190)(180,95)
\path(265,140)(75,90)
\path(265,140)(180,95)
\path(390,185)(370,95)
\path(265,140)(370,95)
\path(265,140)(495,90)(490,90)
\path(470,185)(495,90)
\path(70,70)(160,15)
\path(375,75)(160,15)(160,15)
\path(165,70)(325,15)
\path(495,70)(325,15)(325,15)
\put(190,315){\makebox(0,0)[lb]{\smash{{{\SetFigFont{8}{9.6}{rm}Sign}}}}}
\put(420,237){\makebox(0,0)[lb]{\smash{{{\SetFigFont{8}{9.6}{rm}Rel}}}}}
\put(365,190){\makebox(0,0)[lb]{\smash{{{\SetFigFont{8}{9.6}{rm}Wh-Rel}}}}}
\put(425,190){\makebox(0,0)[lb]{\smash{{{\SetFigFont{8}{9.6}{rm}Non-Wh-Rel}}}}}
\put(315,240){\makebox(0,0)[lb]{\smash{{{\SetFigFont{8}{9.6}{rm}Decl}}}}}
\put(301,190){\makebox(0,0)[lb]{\smash{{{\SetFigFont{8}{9.6}{rm}Y-N}}}}}
\put(275,233){\makebox(0,0)[lb]{\smash{{{\SetFigFont{8}{9.6}{rm}Int}}}}}
\put(245,190){\makebox(0,0)[lb]{\smash{{{\SetFigFont{8}{9.6}{rm}Wh-Int}}}}}
\put(245,145){\makebox(0,0)[lb]{\smash{{{\SetFigFont{8}{9.6}{rm}Headed-Rel}}}}}
\put(130,0){\makebox(0,0)[lb]{\smash{{{\SetFigFont{8}{9.6}{rm}Su-Wh-Rel}}}}}
\put(305,0){\makebox(0,0)[lb]{\smash{{{\SetFigFont{8}{9.6}{rm}That-Rel}}}}}
\end{picture}
}
\end{center}
\caption{HPSG type hierarchy with one-dimensional inheritance}
\label{ale}
\end{figure}

We follow the Carpenter's formalisation of typed feature logic, but
modify the conception of the type hierarchy. In Carpenter's logic
\cite{Carpenter:92},
the type hierarchy is required to be a bounded complete partial
order ({\sc bcpo}), which means that any two types which do
have a common subtype
must have a unique most general common subtype. A type hierarchy
as in figure \ref{hpsg} fails this requirement because the types
{\em Headed-Ph} and {\em Rel} have two common subtypes, but none of them
is more general than the other. In order to make the type
hierarchy a {\sc bcpo}, additional types must be introduced,
resulting in a hierarchy like the one in figure \ref{ale}. In
Carpenter's system, every feature structure has only one unique
most general type, so that it is not possible to assign a feature
structure two types neither of which subsumes the other unless
they have a common subtype.%
\footnote{A similar point holds for the type system of the logic
          programming language {\sc life}
          \cite{Kaci:Podelski:91,Kaci:Lincoln:89}, with the
          difference that the unification of two types need not
          have a unique result;
          the unification of the types {\em Headed-Ph}
          and {\em Rel} would create a choice point and
          produce the two alternative solutions
          {\em Su-Wh-Rel} and {\em That-Rel}.}

Carpenter \cite[pages17-32]{Carpenter:92} describes
a {\em conjunctive type construction} by which
a type hierarchy like the one in figure \ref{hpsg} can be
converted into a bounded complete partial order like in figure
\ref{ale}. Once this is done, efficient algorithms for the
calculation of greatest lower bounds can be used \cite{AitKaci:Nasr:86}.
We argue that such a conjunctive type construction is neither necessary
for theoretical reasons nor for reasons of efficient implementation.

In our system, a feature structure can have different types
as long as they are chosen from different dimensions. Our syntax
for subtype declarations, given in (\ref{eq1}), combines information
about subtyping and disjointness. All the $Y_i$ are subtypes of $X$,
and all $Y_i$ are disjoint.

\begin{equation} X > [Y_1,\ldots,Y_n].
\label{eq1}
\end{equation}

Multi-dimensional inheritance
arises in the case where there is more
than one declaration with the same supertype on the left-hand side,
as in  (\ref{multi}).

\begin{eqnarray} \label{multi}
X & > & [Y_{1.1},...,Y_{1.n}]. \\
\vdots & \vdots   & \vdots       \nonumber \\
X & > & [Y_{m.1},\ldots,Y_{m.k}]. \nonumber
\end{eqnarray}

Instead of writing a separate subtype declaration for each dimension,
multiple dimensions are conventionally connected with the product
operator $*$, as in declaration (\ref{product}), which is equivalent
to the declarations in (\ref{multi}).

\begin{equation}
X >  [Y_{1.1},...,Y_{1.n}] * \ldots * [Y_{m.1},...,Y_{m.k}]. \label{product}
\end{equation}

Multiple inheritance is the case where some $Y$ type occurs
in the right-hand side of more than one type declaration. In this case,
the type has several supertypes, which must be chosen from different
dimensions in order to be consistent with each other.%
\footnote{We ignore the case where one of the supertypes subsumes the
          other since such declarations are redundant.}

We now turn to the semantics of type declarations.
The denotation of each type is a subset of the domain. The semantics
of the type declaration in (\ref{eq1}) is given by the axioms (\ref{subset}),
which states that the denotation of any of the $Y_i$ is a subset of the
denotation of $X$, and (\ref{disjoint}), which states that all the $Y_i$
are disjoint.
No additional axioms are needed in case of multi-dimensional inheritance.

\begin{eqnarray}
   \forall Y_i (\denotation{Y_i} \subseteq \denotation{X}) \label{subset}
\\
   \forall Y_i \forall Y_j (Y_i \neq Y_j \Rightarrow
   \denotation{Y_i} \cap \denotation{Y_j} = \emptyset) \label{disjoint}
\end{eqnarray}

Our system has an open-world semantics for type hierarchies.
A feature structure can be described by two types from different hierarchies,
but there need not be a common subtype of these two types. This
is in contrast with a system like ALE \cite{Carpenter:93} with
a closed-world semantics
where the conjunction of two types is inconsistent
unless one subsumes the other or they have a common subtype.

\begin{figure}[htb]
\begin{verbatim}
sign > [non_headed_ph,headed_ph] * [int,decl,rel].
headed_ph > [h_su,h_co,h_mk,h_fi].
int > [wh_int,y_n].
rel > [wh_rel,non_wh_rel].
h_su > [su_wh_rel].
wh_rel > [su_wh_rel].
h_mk > [that_rel].
non_wh_rel > [that_rel].
\end{verbatim}
\caption{Type declarations}
\label{declarations}
\end{figure}

Our notion of feature typing and appropriateness is based on
Carpenter's feature logic. Every feature is introduced for a
unique most general type, and is appropriate for all subtypes
of that type. In case of multiple inheritance, a type can
inherit different features from its supertypes in different dimensions.
A difference arises with type restrictions for feature values.
In Carpenter's system, the value of a feature has one type as
the type restriction, whereas in our system, the type restriction
can be a conjunction of types from different dimensions.

In our system, the type hierarchy from figure \ref{hpsg} can be expressed
directly with the declarations given in figure \ref{declarations}.

%


Our notion of typing is similar to the one adopted
in the {\em Comprehensive Unification Formalism} ({\sc cuf}
\cite{Doerre:Dorna:93}),
in that it adopts an open-world semantics, and two types are
considered as consistent unless they are explicitly declared to
be disjoint.
However, {\sc cuf} allows to state more general type axioms
using the full power of propositional logic. In our system, the
type axioms are restricted to subtyping (which corresponds to
implication) and disjointness. This restriction allows the
efficient compilation of multi-dimensional inheritance hierarchies
to Prolog terms, which will be described in the next section.

\section{Compilation into a Prolog Term Representation} \label{mellish}

Multi-dimensional type hierarchies have the favourable property that
the types can be compiled to a Prolog term representation. With this
representation, Prolog's built-in term unification is all that is
required to compute the conjunction of two types. The Prolog term
representation given here builds on and extends the representation
introduced by Mellish \cite{Mellish:88,Mellish:92} and used
in the Core Langauge Engine \cite{Alshawi:91} and the ALEP grammar
development system \cite{ET61,ALEP}.

We start out by describing how the translation of
type hierarchies into Prolog terms works, and then give an
example. The translation to terms must be able to handle different
dimensions of typing, mutually exclusive choices in a dimension,
subtyping, multiple inheritance, features, and equality.

\begin{description}
\item [Different dimensions:] Each dimension occupies a different
       argument position in the resulting term representation, so that
information
       from different dimensions can be combined by unification.

\item [Mutually exclusive types:] Mutually exclusive types in the same
      dimension have different functors at the same argument position,
      so that their unification fails.

\item [Subtype:] The term which corresponds to the subtype is a further
      instantiation of the term corresponding to its supertype.

\item [Multiple inheritance:] The term which corresponds to the subtype
       is a further instantiation of the unification of the terms which
       correspond to its supertypes.

\item [Feature:] The term representation has an argument position
       for each feature
       introduced for a type. If a feature is introduced for a subtype,
       then an argument position is provided in that argument which
       further instantiates the supertype.%
\footnote{The compilation of feature structures to Prolog terms has
          been described in \cite{Schoeter:93,Hirsh:86,Covington:89},
          but these works assume untyped feature structures. One may
          wonder what the difference is between features and multiple
          typing dimensions, since they have very similar term
          representations. A technical answer would be that equality
          constraints (coreferences) can be stated over feature values,
          but not over type dimensions. Work by Moshier
          \cite{Moshier:94} on a rational
          reconstruction of typed feature structures in domain theory
          approaches the question from a more fundamental perspective.}
\item [Equality:] In order to be able to distinguish structures that
       are identical from those which just happen to have the same
       value (i.e. their term representation is instantiated to the
       same ground term), an extra variable is introduced in the term
       representation
       (preventing instantiation to a {\em ground} term),
       which is only equal for two structures if they have been made
       identical by unification.
\end{description}

\begin{figure}[htb]
\tt \small
\begin{tabular}{ll}
\mbox{\rm Type} & \mbox{Encoding} \\ \hline
Sign           & \tt \small sign(\_,\_)  \\
Non-Headed-Ph  & \tt \small sign(non\_headed,\_) \\
Headed-Ph      & \tt \small sign(headed\_ph(\_),\_) \\
Int            & \tt \small sign(\_,int(\_)) \\
Decl           & \tt \small sign(\_,decl) \\
Rel            & \tt \small sign(\_,rel(\_)) \\
H-Su           & \tt \small sign(headed\_ph(h\_su(\_)),\_) \\
Wh-Rel         & \tt \small sign(\_,rel(wh\_rel(\_))) \\
Su-Wh-Rel      & \tt \small
           sign(headed\_ph(h\_su(su\_wh\_rel)),rel(wh\_rel(su\_wh\_rel))) \\
...
\end{tabular}
\caption{Encoding of Types as Prolog Terms}
\label{translation}
\end{figure}

Given this kind of representation, two typed feature structures can
be unified simply by unification of the corresponding Prolog terms.
In figure \ref{translation}, we provide an encoding of the types in the
hierarchy from figure \ref{hpsg} by Prolog terms. For simplicity,
we leave out the argument position used to establish the equality
of feature structures, and any argument positions used to encode
features. Note that the type Su-Wh-Rel makes a choice in both dimensions.

If the type Su-Wh-Rel has any subtypes, a choice must be made which of
the two occurences of {\tt su\_wh\_rel} in the Prolog terms should
get argument positions for carrying this information.%
\footnote{From the correctness of the translation, it would be no
          problem representing the subtypes in both occurences, but
          such an encoding is clearly redundant.}
We always choose the leftmost occurence in a term for representing
subtypes (and features). Further occurences then only serve to
make a choice in a particular dimension of the hierarchy, and for that
purpose, an
atom which is distinct from other terms that can occur as alternatives
in the same dimension is sufficient.

\section{Application to Systemic Classification} \label{winograd}

In this section, we apply the conception of multi-dimensional
inheritance to systemic classification networks, which have been
discussed in \cite{Mellish:88}.

\begin{figure}[htb]
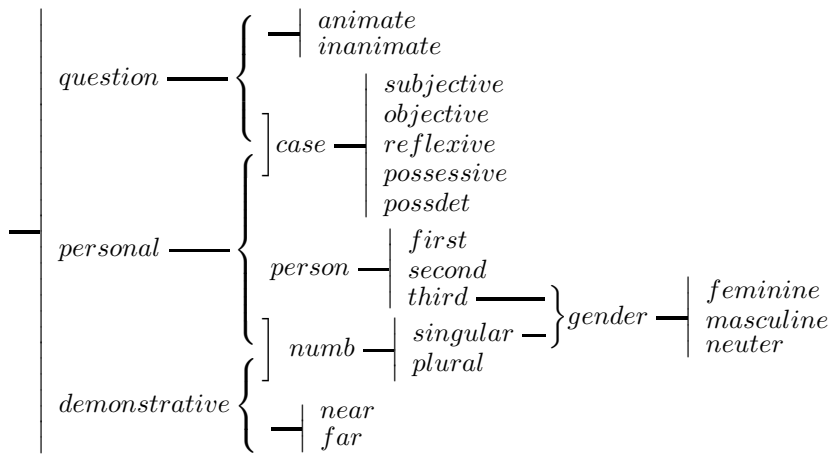

\begin{center}

\newcommand{\choice}{\rule[0.6mm]{4mm}{0.3mm}\hspace*{-1mm}\left|}
\newcommand{\feats}{\left\{}
\newcommand{\mm}{\rule{0mm}{1mm} \\}
\renewcommand{\arraystretch}{0.2}

$
\choice
\begin{array}{l}
   question \ \rule[0.6mm]{8mm}{0.3mm} \feats \begin{array}{l}
            \choice \begin{array}{l}
                          animate \mm  \\
                          inanimate
                          \end{array} \right. \\
             \rule[-5mm]{0mm}{12mm}
             \end{array} \right.

\\
\raisebox{0mm}[0mm][0mm]{$
   \hspace*{27mm} \left] \raisebox{0mm}[3.5mm][3.5mm]{$
                        case \ \choice \begin{array}{l}
                                     subjective \mm \\
                                     objective \mm  \\
                                     reflexive \mm  \\
                                     possessive \mm  \\
                                     possdet
                                     \end{array} \right. $} \right.
$}
\\
   personal \ \rule[0.6mm]{8mm}{0.3mm} \feats
            \begin{array}{l}
             \rule[-5mm]{0mm}{10mm}
            \\
                 person \ \choice \begin{array}{l}
                                 first  \mm \\
                                 second  \mm \\
                                 third \ \rule[0.6mm]{9mm}{0.3mm}
                                 \end{array}
                         \right.
            \\
             \rule[-2.5mm]{0mm}{5mm}
            \end{array}
            \right.
\\
\raisebox{3.5mm}[0mm][0mm]{$
\hspace*{65mm}
             \left\}
             \raisebox{0mm}[3mm][3mm]{$
             gender
             \ \choice \begin{array}{l}
                           feminine  \mm \\
                           masculine \mm  \\
                           neuter
                      \end{array}
             \right.
             $}
 \right.
$}
\\
\raisebox{0mm}[0mm][0mm]{$
   \hspace*{27mm} \left] \begin{array}{l}
                           numb \ \choice \begin{array}{l}
                                    singular\ \rule[0.6mm]{3mm}{0.3mm} \mm  \\
                                    plural
                                       \end{array} \right.
                          \end{array} \right.
$}
\\
   demonstrative \feats \begin{array}{l}
                              \rule[-3mm]{0mm}{6mm}
                         \\
                               \choice \begin{array}{l}
                               near \mm  \\
                               far
                               \end{array} \right.
                         \end{array}
                 \right.
\end{array}
\right.
$

\renewcommand{\arraystretch}{1}

\end{center}
\caption{Winograd's systemic pronoun network}
\label{pronoun}
\end{figure}

Systemic classification networks
are an interesting formalism for such an encoding because they
offer considerable expressive power. Figure \ref{pronoun} shows
a systemic classification network for
English pronouns taken from \cite{Winograd:83}.

Figure \ref{connectives} shows the connectives, and the translation of the
first three in our system. The
final connective (disjunctive entry condition) has no simple
translation. This is not surprising, given the complexity
analysis by Brew \cite{Brew:91} who shows systemic classification
to be NP-hard by giving an encoding of the 3SAT-problem in a
systemic network with disjunctive entry conditions.

We treat disjunctive entry conditions by introduction of new
types into the hierarchy. For each type $X$ at the right-hand side
of a disjunctive entry condition, we introduce two new types,
$X'$ for the original type, and $\neg X$ for its negation. These
pairs of new types are introduced in different dimensions at
the top of the choice system containing the disjunctive
entry conditions. All types $Y$ which have subtypes
that are on the left-hand side of the disjuntive entry condition
for $X$ become subtypes of $X'$, and all other types become
subtypes of $\neg X$. In effect, this is an expansion of the
disjunctive entry conditions to disjunctive normal form.

\begin{figure}[htb]
\begin{center}
\leavevmode\epsfbox{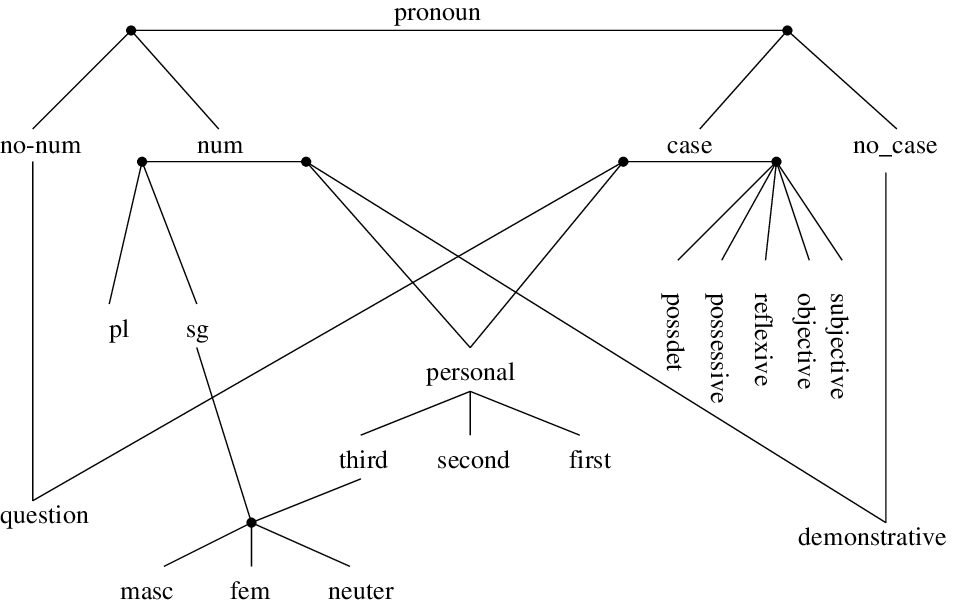}
\end{center}
\caption{Pronoun hierarchy with disjunctions moved to the top}
\label{pronoun2}
\end{figure}

In the worst case, this method can lead to translations which
are exponentially larger than the original classification network,
as for the 3SAT problem.
We show by an example that this need not be the case in practice
by converting the classification of pronouns
given in figure \ref{pronoun} into our system (figure \ref{pronoun2}).

\renewcommand{\arraystretch}{0.5}
\begin{figure}
\begin{tabular}{|l|l|l|}
\hline
\multicolumn{2}{|c|}{Connective} & {\rule[-2mm]{0mm}{5mm} Type Declarataions}
\\ \hline
choice system & \mbox{\rule[-8.5mm]{0mm}{18mm} $ X \left|
                             \begin{array}{l}
                             Y_1 \\ \ \vdots \\ Y_n
                             \end{array} \right. $}  & $X > [Y_1,\ldots,Y_n]$
\\ \hline
multiple choices & \mbox{\rule[-8.5mm]{0mm}{18mm} $ X \left\{
                             \begin{array}{l}
                             Y_1 \\ \ \raisebox{1mm}{\vdots} \\ Y_n
                             \end{array} \right.$}  & $X > Y_1*\ldots*Y_n$
\\  \hline
conjunctive entry cond. & \mbox{\rule[-8.5mm]{0mm}{18mm} $  \left.
                             \begin{array}{l}
                             X_1 \\ \ \vdots \\ X_n
                             \end{array} \right\} Y$}   & $\begin{array}{l}
                                                      X_1 > [Y,\ldots] \\
                                                      \ \vdots \\
                                                      X_n > [Y,\ldots]
                                                      \end{array}$
\\ \hline
disjunctive entry cond. & \mbox{\rule[-8.5mm]{0mm}{18mm} $ \left.
                             \begin{array}{l}
                             X_1 \\ \ \vdots \\ X_n
                             \end{array} \right] Y$}  & $Y' \  \mbox{and}
                                                       \neg Y \ \mbox{move to
                                                       the top of the
                                                       hierarchy}    $
\\ \hline
\end{tabular}
\caption{Systemic network connectives and multi-dimensional inheritance}
\label{connectives}
\end{figure}
\renewcommand{\arraystretch}{1}

In this case, our translation is more efficient than the
``brute-force translation''
described in \cite{Mellish:88} for networks with disjunctive entry conditions.
In Winograd's systemic network, there are 54 possibilities
(6 for interrogative pronouns, 36 for personal pronouns, and
4 for demonstrative pronouns), resulting in a brute-force
translation with 55 arguments.%
\footnote{The ``brute-force translation'' works by encoding $n$ possibilities
          into a term with $n+1$ arguments, whose first and last
          argument are different (either instantiated to different atoms
          or related by an inequality constraint). For encoding a value
          that excludes the $n^{th}$ possilibity, the $n^{th}$ and
          $n+1^{st}$ argument position are unified. If a combination of
          values excludes all possibilities, then all arguments are
          unified with each other, including the first and the last,
          which are different, so that the unification fails.
          There is limited use for this type
          of translation to avoid creation of choice points in
          disjunctions with a small number of possible values. This
          can be regarded as a Prolog implementation of finite domains.}

In contrast, in our type system, the Prolog term translation has at
most 8 nodes in the worst case, as in the following term which
encodes the subjective case masculine (third person singular)
personal pronoun.

\[
\mbox{\tt pronoun(case(personal(third(masc)),subjective),num(personal,sg))}
\]

\section{Conclusion}

We have presented a concept of inheritance which provides direct support
for current linguistic descriptions making use of
``cross-classification'', and can be compiled into an efficient Prolog
term representation.

Given the need for multiple dimensions in lingistic descriptions, we
believe that multi-dimensional type hierarchies will remain important
even when their compilation into Prolog terms is not needed any longer
because unification of typed feature terms will be built-in in future logic
programming languages.

For the time being, however, the combination of multi-dimensional
inheritance and compilation into Prolog terms appears to give both the
efficiency and the expressive power needed to develop larger-scale
grammars and lexicons, and use existing Prolog-based technology
(DCG parsers, left-corner, head-corner, or chart parsers,
semantic-head driven or tabular generators) to build
NLP systems. Such an approach can benefit from all the advantages
of modern Prolog compilers (indexing, coroutining facilities,
module systems etc.) that would need considerable effort to
duplicate in a dedicated grammar formalism.

The multi-dimensional inheritance described in this paper is
implemented in the system ProFIT \cite{Erbach:94}, which translates
programs containing
typed feature terms to ordinary Prolog programs.

\end{document}